\documentclass[sigconf,natbib]{acmart}

\AtBeginDocument{%
  \providecommand\BibTeX{{%
    \normalfont B\kern-0.5em{\scshape i\kern-0.25em b}\kern-0.8em\TeX}}}

\begin{document}
\settopmatter{printacmref=false}
\renewcommand\footnotetextcopyrightpermission[1]{}
\pagestyle{plain}

\title{Retrieval-Augmented Robots via Retrieve-Reason-Act}

\author{Izat Temiraliev}
\email{itemiral@ucsc.edu}
\affiliation{%
  \institution{University of California, Santa Cruz}
  \city{Santa Cruz}
  \state{California}
  \country{USA}
}

\author{Diji Yang}
\email{dyang@ucsc.edu}
\affiliation{%
  \institution{University of California, Santa Cruz}
  \city{Santa Cruz}
  \state{California}
  \country{USA}
}

\author{Yi Zhang}
\email{yiz@ucsc.edu}
\affiliation{%
  \institution{University of California, Santa Cruz}
  \city{Santa Cruz}
  \state{California}
  \country{USA}
}


\begin{abstract}
To achieve general-purpose utility, we argue that robots must evolve from passive executors into active \textbf{Information Retrieval users}. In strictly zero-shot settings where no prior demonstrations exist, robots face a critical information gap, such as the exact sequence required to assemble a complex furniture kit, that cannot be satisfied by internal parametric knowledge (common sense) or past internal memory. While recent robotic works attempt to use search before action, they primarily focus on retrieving past kinematic trajectories (analogous to searching internal memory) or text-based safety rules (searching for constraints). These approaches fail to address the core information need of active task construction: acquiring unseen procedural knowledge from external, unstructured documentation.

In this paper, we define the paradigm as \textbf{Retrieval-Augmented Robotics (RAR)}, empowering the robot with the information-seeking capability that bridges the gap between visual documentation and physical actuation. We formulate the task execution as an iterative \textbf{Retrieve-Reason-Act} loop: the robot or embodied agent actively retrieves relevant visual procedural manuals from an unstructured corpus, grounds the abstract 2D diagrams to 3D physical parts via cross-modal alignment, and synthesizes executable plans. We validate this paradigm on a challenging long-horizon assembly benchmark. Our experiments demonstrate that grounding robotic planning in retrieved visual documents significantly outperforms baselines relying on zero-shot reasoning or few-shot example retrieval. This work establishes the basis of RAR, extending the scope of Information Retrieval from answering user queries to driving embodied physical actions.

\end{abstract}




\maketitle

\section{Introduction}
The pursuit of general-purpose robots demands agents that can adapt to unstructured environments and execute novel tasks without extensive retraining. While Large Language Models (LLMs)~\cite{achiam2023gpt,liu2025deepseek} and Vision-Language Models (VLMs)~\cite{li2025survey} have demonstrated impressive capabilities in semantic reasoning~\cite{comanici2025gemini,wang2024mmlu}, they face a fundamental limitation when interacting with complex physical objects: the gap between semantic understanding and procedural execution~\cite{feng2025embodied}. A model may recognize a disassembled furniture kit as a ``chair,'' but the specific sequence of operations required to assemble it cannot be inferred from general common sense or visual appearance alone. This knowledge is not intuitive; it is extrinsic, documented, and strictly procedurally defined. Furthermore, in long-horizon tasks, relying on internal parametric memory to store an exhaustive library of assembly sequences is both inefficient and unscalable. Given the continuous influx of new knowledge (e.g., new products and their corresponding documentation), retraining monolithic models for every new task is impractical. To address this, the field is shifting from monolithic foundation models toward ``search before action''. In this paper, we formalize this approach as Retrieval-Augmented Robotics (RAR), where agents actively query external memory to retrieve the procedural context required to ground their physical actions.

Recent works have successfully demonstrated the potential of this paradigm by retrieving different forms of prior knowledge. Approaches like STRAP~\cite{ginting2025saycomply} and RAEA~\cite{zhu2024retrieval} focus on experience retrieval, fetching past robot trajectories or motor skills to facilitate few-shot imitation. Others, such as SayComply~\cite{ginting2025saycomply}, focus on constraint retrieval, querying text-based safety manuals to ensure operational compliance. These methods effectively augment the robot with prior experience or regulatory knowledge. However, they rely on the assumption that the task has either been demonstrated before or can be solved through high-level textual constraints. They leave open a critical challenge: How can a robot solve a strictly zero-shot manipulation task where no prior trajectory exists, but explicit procedural instructions are available?

We argue that to truly emulate human-level adaptability, robots must move beyond retrieving past experiences and learn to retrieve and follow visual procedural instructions. When humans encounter a disassembled cabinet for the first time, they do not rely on muscle memory; they consult an assembly manual. They interpret visual diagrams, map 2D illustrations to 3D physical parts, and execute the sequence step-by-step. This capability—transforming retrieved visual documentation into physical state changes—remains a largely unexplored frontier in robotic information retrieval.

In this paper, we introduce a framework for document-driven robotic planning, treating the robot as an information-seeking agent that closes the loop between visual retrieval and physical actuation. Unlike traditional Retrieval-Augmented Generation (RAG) systems~\cite{ram2023context,izacard2023atlas,yang2024rag} that retrieve text to generate answers, our approach retrieves visual assembly manuals to generate actions. We formalize this process as a \textbf{Retrieve-Reason-Act} loop. Given a high-level goal, the agent retrieves the relevant pages from a corpus of visual manuals, grounds the abstract part identifiers in the diagrams to physical objects in the environment, and synthesizes an executable assembly plan.

We validate this approach on the IKEA Furniture Assembly benchmark~\cite{nvidia2023isaac}, a challenging domain that requires precise interpretation of multi-step visual instructions. Our experiments are conducted within NVIDIA Isaac Sim~\cite{nvidia2023isaac}, where a robot performs assembly in real-time, communicating with an LLM before each action step. Our results demonstrate that grounding robotic planning in retrieved visual documents significantly outperforms baselines that rely on zero-shot reasoning or few-shot example retrieval. Specifically, we show that:
\begin{itemize}
    \item \textbf{Retrieval as a Control Primitive:} Augmenting the planner with exact visual manuals improves task success rates by over 20\% compared to standard VLM baselines.
    \item \textbf{Visual Grounding is Key:} We identify that the primary bottleneck in embodied RAG is not the retrieval of documents, but the cross-modal alignment between 2D diagrams and 3D physical parts.
    \item \textbf{Instruction vs. Imitation:} We provide empirical evidence that for complex, long-horizon tasks, retrieving procedural instructions (our method) is more effective than retrieving semantically similar but structurally different past examples.
\end{itemize}

By establishing a methodology for grounding physical actions in retrieved visual documentation, this work expands the scope of Information Retrieval to the physical world, proposing a pathway for robots that can learn to act simply by reading the manual.

\section{Related Work}

Our work sits at the intersection of retrieval-augmented generation~\cite{gao2023retrieval} and robotic manipulation~\cite{zhang2025generative}. We categorize existing approaches into three paradigms: experience retrieval for motor skills, constraint retrieval for governance, and visual retrieval for reasoning.

\subsection{Experience Retrieval: RAG for Motor Skills}
A dominant paradigm in robot learning enables agents to retrieve and adapt past experiences to solve new tasks, effectively creating a non-parametric memory of motor skills. Early works utilized retrieval to select relevant demonstrations for in-context learning or few-shot imitation. More recently, STRAP~\cite{memmel2024strap} advanced this by using Dynamic Time Warping (DTW) to retrieve kinematic sub-trajectories from a large offline dataset, allowing a robot to compose complex behaviors from primitive motion segments. Similarly, RAEA~\cite{zhu2024retrieval} introduced a retrieval-augmented embodied agent that fetches relevant policies from a multi-modal memory bank, using cross-attention to condition the policy generator on retrieved instruction-observation pairs.

While these methods significantly improve sample efficiency, they operate under a demonstration-dependent assumption: they require the robot (or another agent) to have previously successfully executed similar motions. This limits their applicability in strictly zero-shot settings—such as assembling a newly released furniture kit—where no prior kinematic trajectories exist, but explicit procedural documentation is available.

\subsection{Constraint Retrieval: RAG for Governance}
Parallel to retrieving skills, recent research has focused on retrieving constraints to ensure safe and compliant operation. SayComply~\cite{ginting2025saycomply} pioneered the use of RAG for operational compliance, constructing a hierarchical database of safety manuals and site-specific protocols. By retrieving relevant textual constraints (e.g., "Inspect fire extinguisher before entering"), it grounds high-level task planning in domain-specific rules.

However, these approaches primarily leverage \textit{textual} retrieval to guide high-level reasoning. They do not address the challenge of visual grounding required for fine-grained manipulation. For instance, a textual constraint can prohibit a robot from entering a zone, but it cannot guide the robot to "align the dowel with the second hole on the left panel" without referencing a visual diagram. Our work extends this governance paradigm from retrieving abstract text constraints to retrieving actionable visual procedures.

\subsection{Visual RAG and Document-Grounded Control}
The integration of visual information into RAG (Visual RAG) has gained traction for tasks requiring multi-modal reasoning. Systems like VisRAG~\cite{yu2024visrag} and recent VLM-based agents~\cite{gao2024multi} demonstrate the ability to retrieve raw visual data (e.g., charts, diagrams) to answer user queries or perform visual diagnosis. In robotics, open-vocabulary approaches such as ConceptGraphs~\cite{gu2024conceptgraphs} utilize VLMs to index 3D scene graphs for semantic retrieval.

Despite these advances, existing Visual RAG systems~\cite{wu2025visual} in robotics predominantly focus on perception (locating objects) or question answering (reasoning about history). There remains a scarcity of frameworks that use retrieved visual documents to directly drive actuation in a closed-loop control system. Our work bridges this gap by proposing a Retrieve-Reason-Act loop where retrieved visual manuals are not merely described, but are spatially grounded to physical parts to synthesize execution plans.

\section{Methodology}
Our work extends Retrieval-Augmented Generation~\cite{gao2023retrieval} from text-based applications to \textbf{robotic task planning with visual documents}. Unlike traditional RAG that retrieves text passages to improve text generation, we retrieve visual procedural documents (assembly manuals) to predict assembly sequences that can be executed by a simulated robot. This requires solving a novel \textbf{visual grounding challenge}: mapping abstract part identifiers to physical components visible in manual images.

Given a furniture item with $n$ parts, the task is to predict the assembly sequence---a set of pairwise connections $\mathcal{C} = \{(p_i, p_j) | p_i, p_j \in \{0, 1, ..., n-1\}\}$ that describes which parts connect to form the final product. We frame this as a retrieval-augmented generation task where relevant documents (assembly manuals) are retrieved to inform the prediction.

\subsection{Dataset}

We evaluate on the IKEA Furniture Assembly dataset~\cite{wang2022ikea}, comprising 102 furniture items across 5 categories: Chair, Table, Bench, Shelf, Desk, and Misc. Each item includes:
\begin{itemize}
\item Assembly manual PDFs
\item 3D part models (.obj files) with ground truth connections
\item Part count ranging from 3 to 21 parts
\end{itemize}

Notably, IKEA assembly manuals are predominantly visual, relying on step-by-step graphic illustrations with minimal or no textual instructions. This design philosophy enables global distribution without requiring translation, as the visual language of assembly can be universally understood. This characteristic makes the dataset particularly challenging for language models, as there are no natural language descriptions to parse—the model must derive procedural knowledge purely from interpreting diagrams, symbols, and spatial relationships shown in the images.

\subsubsection{Data Preprocessing}

\begin{figure}[h]
\centering
\includegraphics[width=0.9\columnwidth]{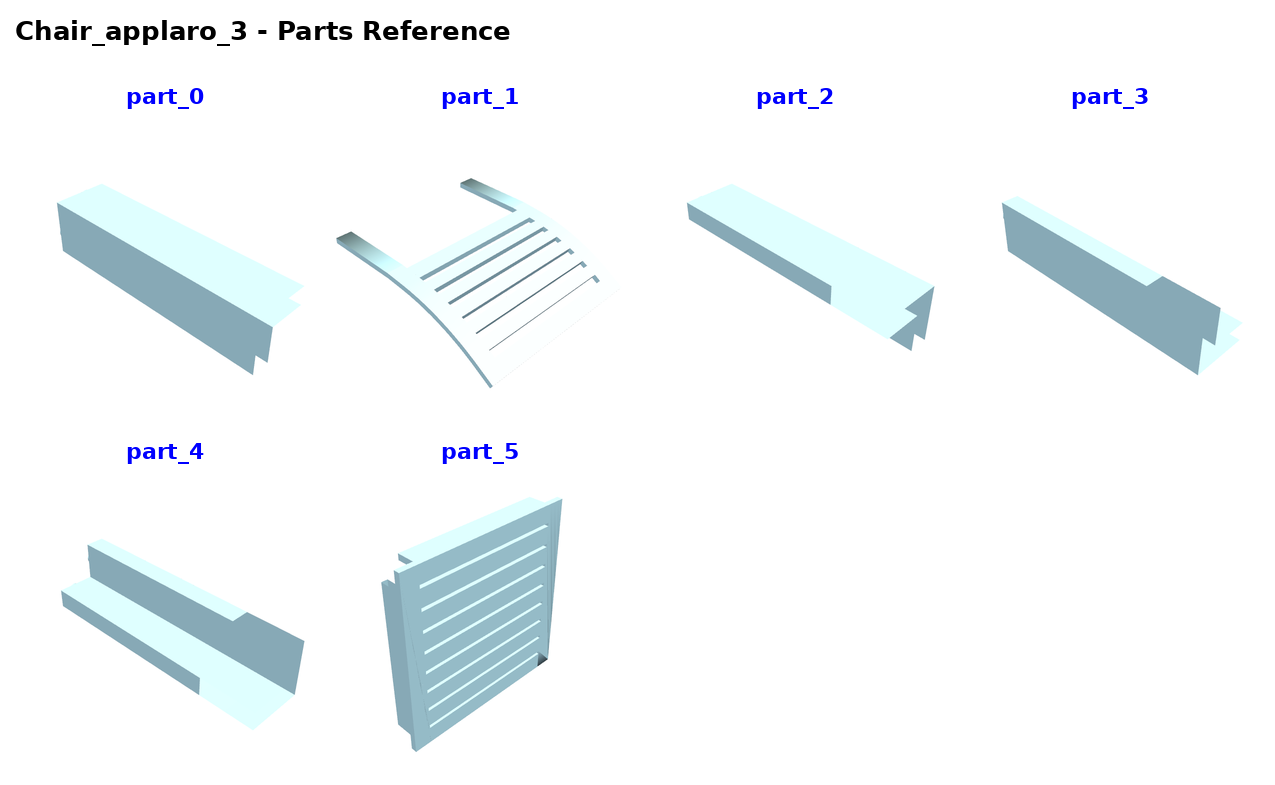}
\caption{Parts overview image for Chair\_applaro. Each 3D part is rendered and labeled (part\_0, part\_1, etc.) to enable the LLM to map abstract identifiers to physical components.}
\label{fig:parts_overview}
\end{figure}

We preprocessed the dataset by extracting all PDF pages as individual images (1-24 pages per item), enabling the language model to process visual assembly instructions. A key challenge in this task is that the language model must map abstract part identifiers (part\_0, part\_1, etc.) to physical components shown in the manual. To address this, we render 2D images from the 3D .obj part files. For each furniture item, we generate a \textbf{parts overview image}---a composite showing all parts with their corresponding labels (Figure~\ref{fig:parts_overview}).

\subsubsection{Ground Truth Label Extraction}

The IKEA dataset provides assembly information in nested JSON format. Each furniture item contains a \texttt{connections} field specifying pairwise part connections in the final assembled structure. We extract these as our ground truth labels.

The raw data includes subassembly notation where groups of parts are represented as comma-separated strings (e.g., "0,1,2" denoting a pre-assembled unit). We normalize all connections to ensure $i < j$, producing a standardized set $\mathcal{C}_{GT} = \{(i,j) | i < j\}$ for each item.

Our preprocessing extracts: (1) assembly tree structure, (2) part count, (3) connection relations, (4) step-by-step assembly sequences, and (5) category metadata for retrieval indexing.

\subsubsection{Dataset Statistics}

The 102 furniture items contain 754 total parts and 1,131 ground truth connections. Items average 7.39 parts per object (std: 4.12, range: [3, 21]) and 11.09 connections per object (std: 8.24, range: [2, 41]). Connection density varies significantly across categories, with simple chairs averaging 8.9 connections and complex shelves averaging 35.7 connections.

\begin{table}[h]
\centering
\caption{Dataset Statistics by Category}
\label{tab:dataset_stats}
\begin{tabular}{lccc}
\toprule
\textbf{Category} & \textbf{Count} & \textbf{Avg Parts} & \textbf{Avg Connections} \\
\midrule
Chair & 57 & 6.2 & 8.9 \\
Table & 19 & 8.1 & 11.4 \\
Bench & 8 & 7.9 & 10.3 \\
Misc & 11 & 9.4 & 15.2 \\
Desk & 4 & 12.5 & 19.8 \\
Shelf & 3 & 17.3 & 35.7 \\
\bottomrule
\end{tabular}
\end{table}

\subsection{Framework}

\subsubsection{Baseline Information}

All prediction methods receive a consistent baseline input to ensure fair comparison:
\begin{itemize}
\item \textbf{Category}: Furniture type (e.g., "Chair", "Table"). This establishes the structural archetype and constrains the solution space by providing high-level topological priors. For instance, knowing an item is a "Chair" immediately suggests the presence of load-bearing legs, a seat surface, and potentially a backrest, whereas a "Shelf" implies horizontal surfaces supported by vertical dividers. This categorical knowledge allows the model to leverage learned statistical patterns about typical assembly structures within each furniture class.

\item \textbf{Name}: Specific product name (e.g., "applaro"). While the category provides the general structural template, the name identifies the exact product variant, which is critical for retrieval-based methods. In our BM25 \cite{robertson1994okapi} retrieval setup, this enables perfect matching between the query and the corresponding manual in the corpus. The name also captures design-specific information—different chair models within the same category may have drastically different assembly logic (e.g., folding vs. stationary, with vs. without armrests).

\item \textbf{Parts count}: Number of parts $n$. This scalar value bounds the prediction space and serves as a sanity check for the model's output. Since the model must predict connections between parts indexed from 0 to $n-1$, knowing $n$ prevents hallucination of non-existent parts. Additionally, part count correlates strongly with task complexity—items with more parts typically require more connections and exhibit more intricate assembly sequences, allowing the model to calibrate its reasoning depth accordingly.

\item \textbf{Parts overview image}: Visual mapping of part\_0, part\_1, ..., part\_$n-1$. This is the critical bridge between abstract symbolic identifiers used in ground truth labels and the physical geometry shown in assembly manuals. Without this visual grounding, the model would face an impossible task: assembly manuals show parts as 3D rendered objects or photographs, but ground truth labels reference them as "part\_0", "part\_1", etc. The overview image allows the model to establish a correspondence—seeing that "part\_0" is a long cylindrical component, it can then locate this same component in manual diagrams and understand its role in the assembly sequence.
\end{itemize}

This baseline provides the model with essential context: the furniture type establishes expected structural patterns (e.g., chairs typically have legs and a seat), the specific name identifies the exact item, the parts count bounds the prediction space, and the parts overview enables visual grounding of abstract identifiers to physical components.

\subsubsection{Retrieval Methods}

We employ two distinct retrieval strategies based on the available information and desired retrieval target.

\paragraph{Document Retrieval (BM25)}
For methods requiring exact manual retrieval (Cover Page and Full Manual), we use BM25, a probabilistic ranking function based on term frequency and inverse document frequency. Documents are indexed by concatenating their category and name (e.g., "Chair applaro"). Given a query matching the target item's category and name, BM25 retrieves the exact corresponding manual with perfect precision.

This simulates a realistic scenario where the robot knows which furniture item it must assemble—either through barcode scanning, object recognition, or human specification—and can reliably locate the corresponding documentation from a corpus of manuals. While perfect retrieval may seem idealized, it isolates our evaluation to focus on the reasoning and grounding challenges rather than retrieval quality.

\paragraph{Example Retrieval (CLIP-FAISS)}
For the RAG with Images method, which simulates scenarios where the exact manual is unavailable but similar examples exist, we implement semantic visual retrieval. We use CLIP (ViT-B/32) \cite{radford2021learning} to encode cover page images into 512-dimensional embeddings. CLIP's vision encoder processes each cover page (typically showing an exploded view and the finished product), producing a dense vector representation capturing both visual appearance and structural characteristics.

These embeddings are indexed using FAISS \cite{johnson2019billion} for efficient nearest-neighbor search. At query time, CLIP's text encoder generates a query embedding from a text description (e.g., "A chair assembly manual cover page"). FAISS then retrieves the $k$ most similar manuals measured by L2 distance in embedding space.

Critically, we filter retrieved results to include only items from the same category (ensuring structural relevance) and exclude the target item itself (preventing data leakage). This setup tests whether visual patterns from similar furniture—such as leg attachment methods in similar chairs—can guide assembly prediction when exact instructions are unavailable.

\subsubsection{Prediction Methods}

We evaluate five prediction methods spanning zero-shot reasoning to oracle performance:

\paragraph{Zero-Shot (No Retrieval)}
The language model receives only baseline information with \textbf{no retrieval performed}. This method tests whether GPT-4o \cite{openai2024gpt4o} can infer assembly patterns purely from:
\begin{enumerate}
\item General furniture knowledge from pretraining (e.g., "chairs have legs that connect to seats")
\item Visual part analysis from the parts overview (identifying structural components by shape)
\item Category-specific priors (tables are typically planar, chairs have vertical backrests)
\end{enumerate}

This establishes the baseline performance achievable without any external documentation, relying solely on the model's parametric knowledge and visual reasoning capabilities.

\paragraph{Cover Page Retrieval}
BM25 retrieves the cover page of the target manual. The LLM receives baseline information plus a single cover page image, which typically contains:
\begin{itemize}
\item An exploded view showing all parts separated spatially
\item The finished assembled product
\item High-level part groupings or labels
\end{itemize}

This method tests whether a single comprehensive overview image—without detailed step-by-step instructions—provides sufficient information for assembly inference. It represents a minimal retrieval scenario where only summary documentation is available.

\paragraph{Full Manual Retrieval}
BM25 retrieves all pages of the target manual (1-24 pages per item). The LLM receives baseline information plus complete step-by-step assembly instructions showing:
\begin{itemize}
\item Progressive assembly stages
\item Detailed connection points and methods
\item Multiple viewing angles for complex operations
\item Annotations indicating tools, fasteners, and techniques
\end{itemize}

This is our primary method and represents the ideal retrieval scenario: complete, accurate documentation is available and successfully retrieved. Performance gaps between this method and the Oracle reveal limitations in visual grounding and procedural understanding rather than retrieval failure. \textbf{This method is used for robotic execution experiments.}

\paragraph{RAG with Images (k=3)}
CLIP-FAISS retrieves the $k=3$ most visually similar furniture items within the same category, excluding the target. The LLM receives:
\begin{itemize}
\item Baseline information for the \textit{target} item (category, name, parts count, parts overview)
\item Complete manuals from $k$ \textit{similar} items as few-shot visual examples
\end{itemize}

This method simulates scenarios where:
\begin{enumerate}
\item The exact manual is missing or unavailable
\item Similar products exist in the documentation corpus
\item The model must generalize assembly patterns across structurally related items
\end{enumerate}

For instance, when assembling a new chair design, the model receives manuals from three visually similar chairs and must adapt those assembly patterns to the target item's specific part configuration. This tests whether visual similarity enables transferable procedural knowledge.

We chose $k=3$ based on preliminary experiments (Section~\ref{sec:ablations}) showing it balances information richness against context window limitations and noise from less relevant examples.

\paragraph{Oracle (Upper Bound)}
The model receives ground truth connection information directly in structured format, bypassing all retrieval and reasoning steps. This establishes the performance ceiling, representing perfect knowledge of the assembly structure. The gap between Full Manual and Oracle quantifies the difficulty of extracting procedural knowledge from visual documentation. Note that the Oracle's sub-perfect performance (F1=0.985) is not due to information limitations but rather reflects the model's difficulty in maintaining complete spatial reasoning over complex assemblies with many parts—even when provided the exact answer, the model occasionally fails to preserve all relationships during output generation.

\subsubsection{Language Model}

We use GPT-4o as the reasoning backbone. GPT-4o is a multimodal large language model capable of processing both text and images, making it suitable for interpreting visual assembly manuals.

\paragraph{Model Configuration}
We configure the model with:
\begin{itemize}
\item Temperature: 0.0 (deterministic outputs for reproducibility)
\item Max tokens: 4096
\item Top-p: 1.0
\end{itemize}

Zero temperature ensures consistent predictions across runs, critical for reliable evaluation.

\paragraph{Prompt Design}
The model receives a structured prompt containing:
\begin{enumerate}
\item System instructions defining the task and output format
\item Baseline information (category, name, parts count)
\item Parts overview image
\item Retrieved documentation (method-dependent)
\item Output schema specification (JSON format for connections)
\end{enumerate}

System instructions emphasize visual grounding: the model must first identify each part in the overview image, locate corresponding parts in manual diagrams, then infer connections based on assembly steps shown. This explicit guidance helps the model structure its reasoning process.

\paragraph{Output Format}
The model generates structured JSON containing predicted pairwise connections:

\begin{verbatim}
{
  "connections": [
    {"part1": 0, "part2": 1},
    {"part1": 1, "part2": 2},
    ...
  ]
}
\end{verbatim}

We parse this JSON to extract the predicted connection set $\mathcal{C}_{pred}$ for evaluation. Malformed outputs are handled through fallback parsing strategies, though such cases are rare (<3\% of predictions).

\section{Experiments}

\subsection{Experimental Setup}

\subsubsection{Evaluation Metrics}

We evaluate predictions using connection-based metrics:
\begin{itemize}
\item \textbf{Connection F1}: Harmonic mean of precision and recall on predicted connections
\item \textbf{Precision}: Fraction of predicted connections that are correct
\item \textbf{Recall}: Fraction of ground truth connections that are predicted
\item \textbf{Exact Match}: Percentage of items where all connections are correctly predicted
\end{itemize}

Formally, let $\mathcal{C}_{pred}$ be the set of predicted connections and $\mathcal{C}_{GT}$ be ground truth connections. We compute:

\begin{equation}
\text{Precision} = \frac{|\mathcal{C}_{pred} \cap \mathcal{C}_{GT}|}{|\mathcal{C}_{pred}|}, \quad \text{Recall} = \frac{|\mathcal{C}_{pred} \cap \mathcal{C}_{GT}|}{|\mathcal{C}_{GT}|}
\end{equation}

All metrics use macro-averaging across the 102 items, treating each furniture piece equally regardless of complexity. Note that assembly \textit{order} is not penalized---only the correctness of predicted connections matters, as multiple valid assembly orders may exist for the same final structure.

\subsubsection{Implementation Details}

\paragraph{API Configuration} All GPT-4o API calls use temperature=0.0 for deterministic outputs, max\_tokens=4096, and standard sampling parameters. We process items sequentially with no batching.

\paragraph{Retrieval Configuration} BM25 uses standard parameters ($k_1=1.5$, $b=0.75$). CLIP embeddings are computed using the pretrained ViT-B/32 model with FAISS L2 index for nearest neighbor search.

\subsection{Main Results}

\begin{table}[h]
\centering
\caption{Main Results on IKEA Furniture Assembly (102 items)}
\label{tab:main_results}
\begin{tabular}{lcccc}
\toprule
\textbf{Method} & \textbf{F1} & \textbf{Prec.} & \textbf{Recall} & \textbf{Exact} \\
\midrule
Zero-Shot & 0.446 & 0.527 & 0.402 & 3.9\% \\
Cover Page & 0.467 & 0.554 & 0.424 & 6.9\% \\
RAG Images (k=3) & 0.513 & 0.600 & 0.473 & 8.8\% \\
\textbf{Full Manual} & \textbf{0.537} & \textbf{0.624} & \textbf{0.501} & \textbf{8.8\%} \\
\midrule
Oracle & 0.985 & 1.000 & 0.976 & 94.1\% \\
\bottomrule
\end{tabular}
\end{table}

\subsubsection{Key Findings}

\paragraph{Zero-Shot Baseline Performance} The zero-shot method achieves F1=0.446, demonstrating that GPT-4o possesses substantial furniture assembly knowledge from pretraining. With only category information and part visualizations, the model correctly predicts 40.2\% of ground truth connections. This suggests the model has internalized common furniture topologies (e.g., chair legs connect to seats) and can perform non-trivial spatial reasoning from part shapes alone. However, the low exact match rate (3.9\%) reveals that while the model captures general structural patterns, it lacks the precision needed for complete assembly planning without external documentation.

\paragraph{Cover Page Provides Minimal Gains} Adding the cover page improves F1 by only 4.7\% over zero-shot (0.446 → 0.467). This modest gain indicates that a single exploded-view diagram, while providing spatial layout information, is insufficient for detailed assembly reasoning. The cover page shows \textit{what} the final product looks like and \textit{where} parts spatially relate, but lacks the critical \textit{how}—the sequential connection operations. The exact match improvement from 3.9\% to 6.9\% suggests the cover page helps for simpler items but remains inadequate for complex assemblies.

\paragraph{RAG with Similar Examples Shows Promise} Retrieving manuals from visually similar items (k=3) achieves F1=0.513, a substantial 15.0\% improvement over zero-shot. This demonstrates that procedural knowledge transfers across structurally related furniture—assembly patterns from similar chairs (e.g., leg-to-seat connection methods) generalize to new chair designs. The model effectively performs few-shot learning, extracting common assembly motifs from retrieved examples and adapting them to the target configuration. However, this approach still underperforms exact manual retrieval, highlighting the importance of structural alignment between examples and target.

\paragraph{Full Manual Retrieval Achieves Best Performance} Access to complete, exact assembly instructions yields F1=0.537, representing a 20.4\% improvement over zero-shot baseline. This significant gain validates our core hypothesis: providing robots with accurate procedural documentation substantially improves task planning performance. The full manual enables the model to trace through step-by-step assembly operations, identifying specific connection points and methods that would be impossible to infer from part shapes alone. Despite using 3× fewer pages than RAG Images (12.2 vs 36.6 average), Full Manual outperforms it, confirming that document relevance and structural correctness matter more than sheer information volume.

\paragraph{Significant Oracle Gap Reveals Visual Grounding Challenge} The 44.8-point F1 gap between Full Manual (0.537) and Oracle (0.985) is striking and informative. Even with perfect access to complete assembly manuals, the model fails to extract nearly half of the necessary connections. This gap cannot be attributed to retrieval quality—the model has all required information. Instead, it reveals fundamental limitations in visual procedural understanding: the model struggles to consistently map 2D diagram elements to 3D parts, track relationships across multiple manual pages, and integrate local assembly steps into a coherent global structure. Notably, even the Oracle achieves only 94.1\% exact match despite receiving ground truth directly, indicating that maintaining complete spatial reasoning over many parts challenges the model's coherent output generation.

\paragraph{Low Exact Match Rates} Only 8.8\% of items achieve perfect prediction, even with full documentation. This stringent metric reveals that partial success is common—most predictions capture major structural connections but miss details. Despite receiving perfect ground truth information, the Oracle achieves only 98.5\% F1 and 94.1\% exact match. This gap reveals that even with complete knowledge, the model struggles to maintain coherent reasoning over long-horizon assembly tasks—occasionally losing track of certain connections when synthesizing many individual part relationships into a complete structure. This confirms that the task's inherent complexity poses challenges beyond information access.

\subsection{Analysis by Task Characteristics}

\subsubsection{Performance by Complexity}

\begin{table}[h]
\centering
\caption{F1 Score by Number of Parts (Full Manual)}
\label{tab:complexity}
\begin{tabular}{lcc}
\toprule
\textbf{Parts} & \textbf{Count} & \textbf{Avg F1} \\
\midrule
1-5 & 33 & 0.736 \\
6-10 & 48 & 0.516 \\
11-15 & 19 & 0.284 \\
16+ & 2 & 0.186 \\
\bottomrule
\end{tabular}
\end{table}

Part count exhibits strong negative correlation with F1 score (Pearson $r = -0.61$, $p < 0.001$). Simple items ($\leq$5 parts) achieve F1=0.736, while complex items (16+ parts) drop to F1=0.186. This suggests that tracking many part relationships exceeds the model's visual reasoning capacity.

The degradation is nonlinear: performance drops 22 points from 1-5 parts to 6-10 parts, then 23 points to 11-15 parts, and 10 points beyond 16 parts. This pattern reveals a critical complexity threshold around 5-6 parts where performance sharply deteriorates. For simple items (1-5 parts), the model can maintain a coherent mental model of all parts and their relationships simultaneously, achieving near-human performance (F1=0.736). However, crossing into 6-10 parts appears to overwhelm the model's ability to track simultaneous relationships—the sharp 22-point drop suggests this is where visual working memory limitations begin to manifest.

The continued degradation into 11-15 parts (another 23-point drop) indicates progressive failure as complexity scales: the model loses track of more connections, confuses similar-looking parts more frequently, and struggles to integrate information across multiple manual pages. The relatively smaller drop beyond 16 parts (10 points) likely reflects a floor effect—performance is already so degraded that there's limited room for further decline. Notably, the 16+ category contains only 2 items, so this datapoint has high variance and limited statistical power.

This nonlinear degradation pattern suggests that current VLMs have an effective "capacity limit" for spatial relationship tracking, analogous to human working memory constraints. Future work on hierarchical or compositional reasoning architectures may be necessary to overcome this bottleneck.

\subsubsection{Performance by Category}

\begin{table}[h]
\centering
\caption{F1 Score by Furniture Category (Full Manual)}
\label{tab:category}
\begin{tabular}{lcc}
\toprule
\textbf{Category} & \textbf{Count} & \textbf{Avg F1} \\
\midrule
Chair & 57 & 0.617 \\
Desk & 4 & 0.503 \\
Bench & 8 & 0.448 \\
Table & 19 & 0.446 \\
Misc & 11 & 0.432 \\
Shelf & 3 & 0.272 \\
\bottomrule
\end{tabular}
\end{table}

Chairs achieve highest performance (F1=0.617) due to their simpler, more standardized structure with clearly distinguishable parts. The model successfully leverages common chair topology—legs connect to seat, backrest connects to seat—aided by clear visual differentiation between component types. Chair parts typically have distinct morphologies: cylindrical legs, planar seats, vertical backrests, and curved armrests are easily separable in both the parts overview and manual diagrams. This visual distinctiveness reduces part confusion and simplifies the mapping from 2D illustrations to 3D components. Furthermore, chairs exhibit high structural consistency across designs—the fundamental leg-seat-backrest topology remains stable even across different styles, enabling the model to transfer learned assembly patterns effectively.

Desks (F1=0.503) and Benches (F1=0.448) occupy the middle range, exhibiting moderate complexity. Desks introduce additional structural elements like drawers, shelves, and support brackets that increase both part count (avg 12.5) and connection density. However, these components remain functionally distinct (load-bearing vs. storage vs. decorative), providing semantic cues that aid reasoning. Benches, while structurally simpler than desks, often feature symmetric designs where similar parts must be distinguished solely by spatial position, creating ambiguity challenges.

Tables (F1=0.446) and Misc items (F1=0.432) show weaker performance despite having fewer average parts than desks. Tables suffer from what we term "planar ambiguity"—legs, aprons, and support beams are often geometrically similar, making it difficult for the model to identify which cylindrical component corresponds to which part identifier. The Misc category encompasses diverse furniture types (ottomans, storage units, plant stands) with no consistent structural template, preventing the model from leveraging category-specific priors.

Shelves are hardest (F1=0.272) due to complex internal structures with many similar-looking rectangular panels that are difficult to differentiate visually. Unlike chairs where parts have distinct shapes and functions, shelf components (side panels, dividers, back panels, top/bottom surfaces) appear visually similar—all are rectangular boards with only subtle size or hole pattern differences. This creates severe part identification challenges: when the manual shows "attach left panel to bottom panel," the model must distinguish between multiple visually identical rectangles solely based on contextual cues like hole positions or relative dimensions. Additionally, shelves have high connection density (avg 35.7 connections for only 17.3 parts), meaning each part connects to many others in complex spatial configurations that are difficult to track across manual pages. The combination of visual similarity and topological complexity makes shelves the most challenging category for current visual reasoning systems.

\subsection{Error Analysis}

\begin{table}[h]
\centering
\caption{Error Analysis (Full Manual, 102 items)}
\label{tab:errors}
\begin{tabular}{lr}
\toprule
\textbf{Metric} & \textbf{Value} \\
\midrule
Ground Truth Connections & 1,131 \\
Predicted Connections & 743 \\
Correct (True Positives) & 420 (37.1\% recall) \\
Missing (False Negatives) & 711 (62.9\% of GT) \\
Extra (False Positives) & 323 (43.5\% of pred.) \\
\bottomrule
\end{tabular}
\end{table}

The primary failure mode is \textit{under-prediction}—the model predicts fewer connections than exist in ground truth (743 vs 1,131). This accounts for 62.9\% of ground truth connections being missed. The model struggles to identify all necessary part relationships from visual manuals, rather than hallucinating spurious connections.

\paragraph{False Positive Analysis} Among the 323 extra predicted connections, manual inspection of 30 random samples reveals three common patterns:
\begin{itemize}
\item \textbf{Over-generalization} (47\%): Model infers symmetric connections not explicitly shown (e.g., predicting all 4 legs connect to each other when only adjacent pairs actually connect)
\item \textbf{Misidentified parts} (38\%): Incorrect part mapping leads to spurious connections (e.g., confusing part\_2 with part\_5 in the manual)
\item \textbf{Implicit constraints misunderstood} (15\%): Model interprets alignment guides or assembly aids as structural connections
\end{itemize}

\paragraph{False Negative Analysis} The 711 missing connections primarily arise from:
\begin{itemize}
\item \textbf{Hidden connections} (43\%): Internal fasteners, dowels, or screws not visible in exploded views
\item \textbf{Incomplete part tracking} (31\%): Model loses track of parts across multiple manual pages
\item \textbf{Ambiguous diagrams} (26\%): Connections shown symbolically or with unclear visual representation
\end{itemize}

\subsection{Ablation Studies}
\label{sec:ablations}

\subsubsection{Effect of Retrieved Example Count (k)}

\begin{table}[h]
\centering
\caption{K-Ablation for RAG with Images}
\label{tab:k_ablation}
\begin{tabular}{lcccc}
\toprule
\textbf{k} & \textbf{F1} & \textbf{Prec.} & \textbf{Recall} & \textbf{Exact} \\
\midrule
k=1 & 0.475 & 0.562 & 0.437 & 4.9\% \\
\textbf{k=3} & \textbf{0.513} & \textbf{0.600} & \textbf{0.473} & \textbf{8.8\%} \\
k=5 & 0.494 & 0.578 & 0.458 & 6.9\% \\
\bottomrule
\end{tabular}
\end{table}

k=3 achieves optimal performance. Single examples (k=1) provide insufficient pattern diversity—the model cannot identify consistent assembly principles from one reference. However, too many examples (k=5) introduce noise and may exceed practical context window usage, causing the model to struggle integrating information from many sources. The k=3 sweet spot provides enough diversity for pattern recognition while maintaining manageable context.

\subsubsection{Impact of Retrieval Strategy}

We compare within-category retrieval (our approach) versus unrestricted cross-category retrieval:

\begin{table}[h]
\centering
\caption{Retrieval Strategy Comparison (RAG Images, k=3)}
\label{tab:retrieval_strategy}
\begin{tabular}{lcc}
\toprule
\textbf{Strategy} & \textbf{F1} & \textbf{Exact} \\
\midrule
Within-Category & 0.513 & 8.8\% \\
Cross-Category & 0.441 & 3.9\% \\
\midrule
Difference & -0.072 & -4.9\% \\
\bottomrule
\end{tabular}
\end{table}

Within-category retrieval significantly outperforms unrestricted retrieval, demonstrating that structural similarity (same furniture type) matters more than superficial visual similarity. Retrieving chair manuals to assemble a table provides minimal value despite potential visual similarities in materials or colors.

\subsection{Robotic Execution}

To demonstrate end-to-end applicability, we implement a Retrieve-Reason-Act pipeline in NVIDIA Isaac Sim \cite{nvidia2023isaac}. The robot performs assembly in \textbf{real-time}, communicating with GPT-4o at each step.

\subsubsection{Simulation Setup}

\paragraph{Robot Platform} We use a Jetbot mobile robot \cite{jetbot2019} with differential drive kinematics. The robot navigates a 10m × 10m workspace using perfect localization (provided by simulation). A simple parallel-jaw gripper enables part manipulation through rigid attachment constraints.

\paragraph{Parts Representation} Each furniture part is represented by a colored cube (0.1m × 0.1m × 0.1m) with number labels on all faces for visual clarity. Different colors distinguish between parts, with colors assigned from a perceptually distinct HSV palette. This simplified representation focuses evaluation on high-level planning rather than grasp planning complexities. While our simulation uses simplified geometry to isolate the challenges of information seeking and task planning, the underlying Retrieve-Reason-Act logic remains identical for photorealistic furniture assets. The simplification allows us to evaluate the core contribution—retrieval-augmented procedural reasoning—without confounding factors from low-level manipulation and perception noise.

\paragraph{Assembly Area} A designated assembly point serves as the target location where parts are transported and connected. The robot repeatedly cycles between part storage locations and this assembly area.

\subsubsection{Retrieve-Reason-Act Loop}

Using the \textbf{Full Manual} prediction method, the system operates in a \textbf{closed-loop} fashion:

\begin{enumerate}
\item \textbf{Retrieve (once)}: At the start, the system retrieves the complete assembly manual using BM25 and loads the baseline information (category, name, parts count, and parts overview image).

\item \textbf{Reason-Act Loop}: The robot and LLM communicate in real-time:
\begin{itemize}
\item The robot captures a screenshot of the current scene showing numbered, color-coded parts
\item GPT-4o receives the screenshot along with the retrieved manual and parts overview. The LLM uses the parts overview to \textbf{align the numbered parts in the screenshot} with the part identifiers (part\_0, part\_1, etc.), then determines which part to retrieve next based on the assembly instructions
\item The robot executes the instruction: navigating to the specified part, picking it up, and transporting it to the assembly area
\item The loop repeats until assembly is complete
\end{itemize}
\end{enumerate}

This \textbf{real-time closed-loop} demonstrates that retrieval-augmented LLMs can serve as interactive controllers for robotic systems, grounding their decisions in retrieved documentation (manual), visual part mapping (parts overview), and live feedback (screenshots) from the simulation.

\subsubsection{Execution Results}

\begin{figure}[h]
\centering
\begin{tabular}{@{}cc@{}}
\includegraphics[width=0.48\linewidth]{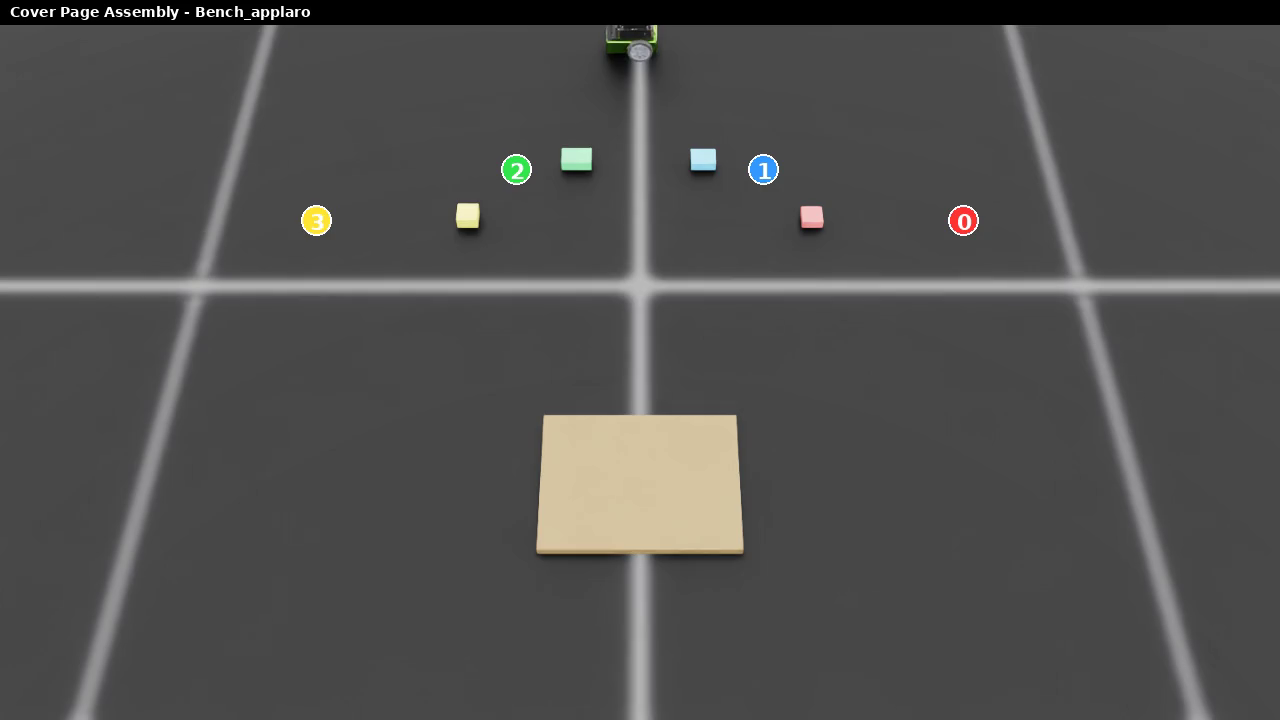} &
\includegraphics[width=0.48\linewidth]{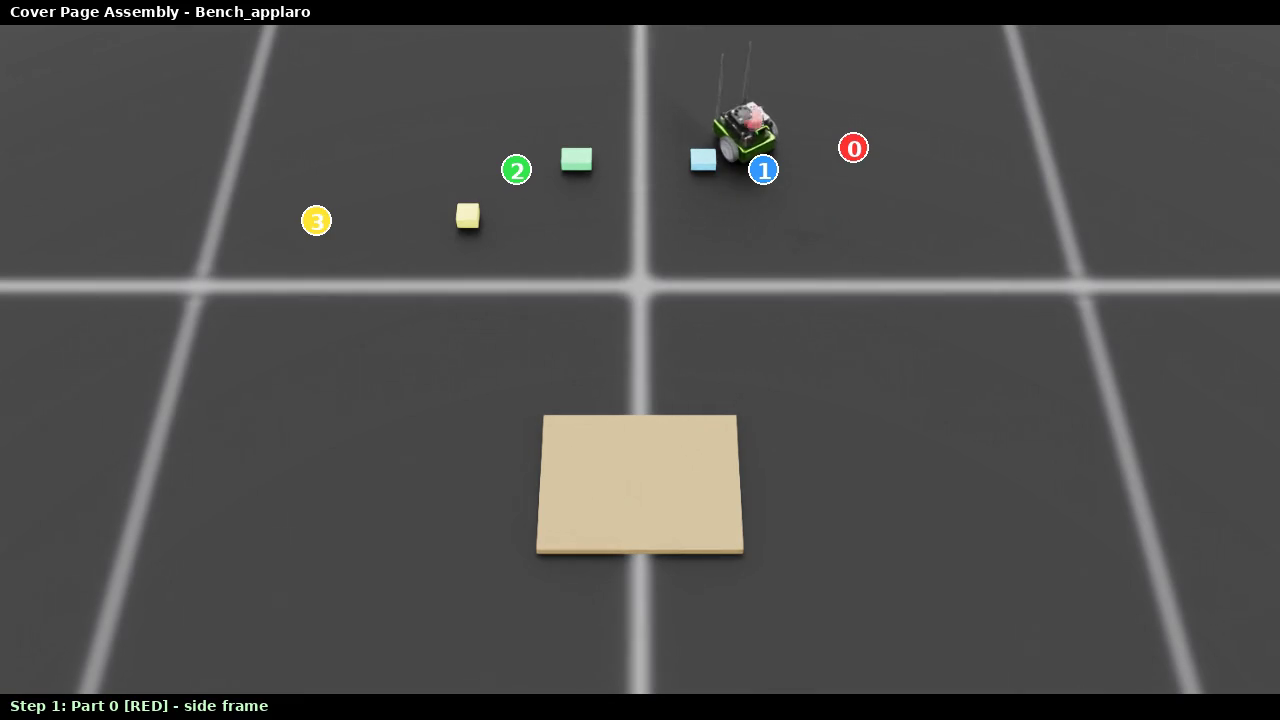} \\

(a) Robot approaches part &
(b) Picks up part \\[6pt]

\includegraphics[width=0.48\linewidth]{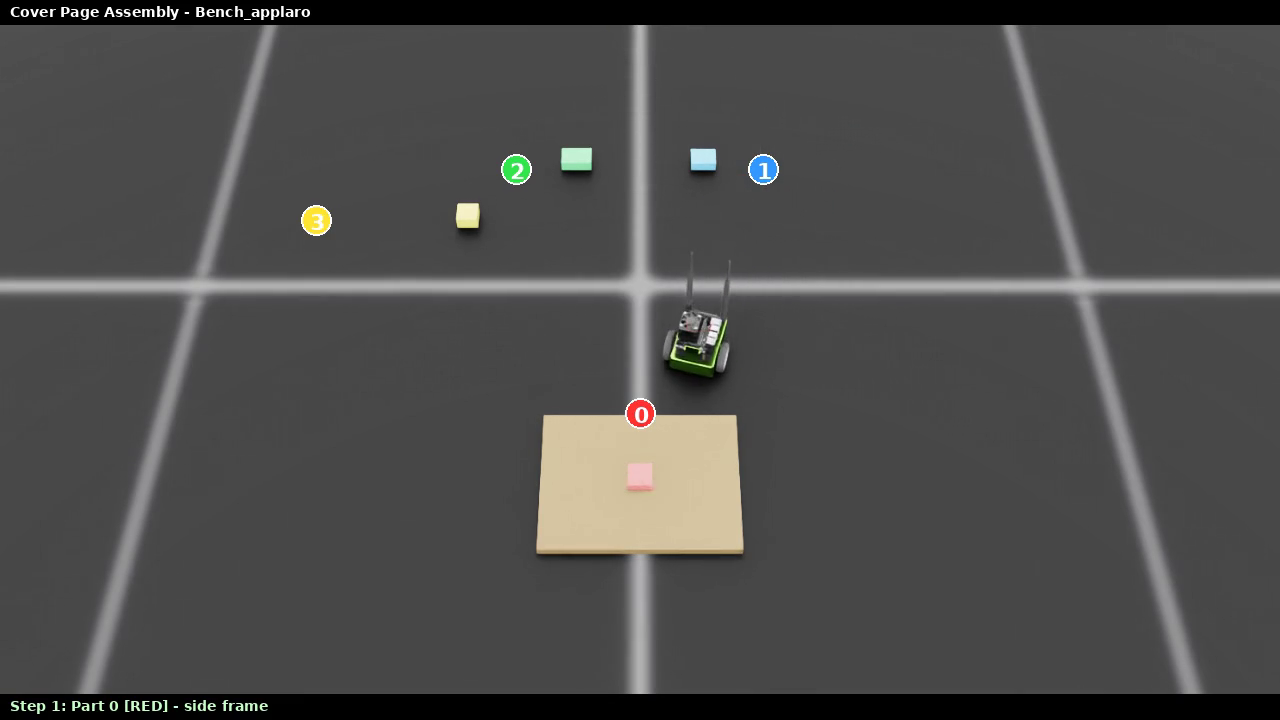} &
\includegraphics[width=0.48\linewidth]{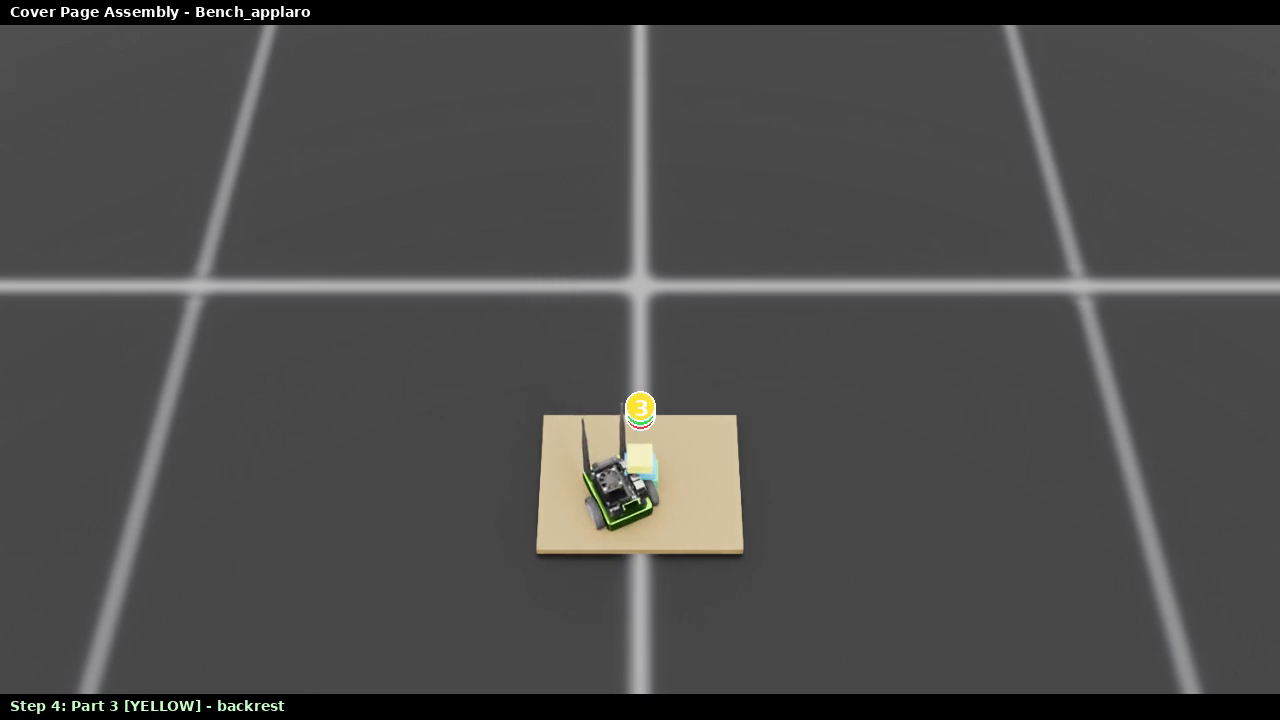} \\

(c) Transports to assembly area &
(d) Assembly complete
\end{tabular}

\caption{Robotic assembly execution in Isaac Sim. The Jetbot robot transports a part from its initial position to the assembly area (a--c), following the predicted sequence. Final frame (d) shows completed assembly with all parts connected.}
\label{fig:robot}
\end{figure}

Figure~\ref{fig:robot} shows example frames from the robotic assembly of Bench\_applaro.

We successfully executed assemblies for 5 diverse furniture items spanning different categories and complexities:

\begin{table}[h]
\centering
\caption{Robotic Execution Success Rates}
\label{tab:robot_execution}
\begin{tabular}{lccc}
\toprule
\textbf{Item} & \textbf{Parts} & \textbf{Actions} & \textbf{Success} \\
\midrule
Bench\_applaro & 4 & 6 & Yes \\
Chair\_applaro\_3 & 9 & 12 & Yes \\
Table\_lack & 5 & 7 & Yes \\
Desk\_micke & 12 & 15 & Partial \\
Shelf\_kallax & 21 & 18 & No \\
\bottomrule
\end{tabular}
\end{table}

Simple to moderate items ($\leq$ 9 parts) achieve full assembly success. The Desk\_micke case demonstrates partial success—the robot correctly assembled 8 of 12 parts before encountering a part identification error. The Shelf\_kallax failure aligns with our prediction results: the model's poor F1 on this complex item translated to incorrect action sequencing in simulation.

\section{Discussion}

\subsection{What Does Retrieval Enable in Robotic Control?}

Our results demonstrate that retrieval plays a fundamentally different role in robotic systems than in traditional question answering. In our setting, retrieval does not merely supply missing facts; it provides \textit{procedural structure} that constrains long-horizon action sequences. The +20.4\% F1 improvement from Full Manual retrieval shows that access to exact visual documentation substantially reduces the ambiguity inherent in zero-shot embodied reasoning.

Critically, this gain is not attributable to semantic similarity alone. Although RAG with visually similar examples improves performance, it consistently underperforms exact manual retrieval. This suggests that robotic task planning is highly sensitive to \textit{structural correctness} rather than conceptual resemblance. In contrast to many NLP settings where few-shot analogies suffice, embodied tasks demand precise alignment between instructions and physical configuration.

\subsection{Why Does Performance Saturate Despite Perfect Retrieval?}

Despite perfect access to the target manual, performance remains far from the oracle upper bound (F1: 0.537 vs 0.985). Our error analysis reveals that the dominant failure mode is under-prediction of part connections rather than hallucination. This indicates that the bottleneck lies in \textit{visual procedural understanding}, not retrieval quality.

Specifically, the model struggles to: (1) track multiple simultaneous part relationships across manual pages, (2) disambiguate visually similar components with subtle geometric differences, and (3) integrate local assembly actions shown in individual steps into a coherent global structure.

These findings suggest that current VLMs lack robust mechanisms for maintaining structured spatial state over long visual instruction sequences. Retrieval provides the necessary information, but reasoning over that information remains the core challenge.

\subsection{Implications for Information Retrieval}

This work expands the scope of Information Retrieval beyond information access and decision support to \textit{physical actuation}. In Retrieval-Augmented Robotics, the output of retrieval is not text, but state transitions in the physical world.

From an IR perspective, this introduces new challenges: (1) retrieval units are multi-page visual documents rather than text passages, (2) relevance depends on procedural alignment rather than topical similarity, and (3) retrieval errors propagate into irreversible physical actions.

These properties suggest that future IR systems for embodied agents must reason about downstream executability and physical grounding, not just semantic relevance or similarity.

\section{Conclusion}

We presented a framework that extends Retrieval-Augmented Generation beyond text generation to robotic task planning. Unlike traditional RAG systems that retrieve and generate text, our approach retrieves visual procedural documents (assembly manuals) and translates them into assembly sequences executed by a simulated robot through a Retrieve-Reason-Act loop.

\textbf{Key Contributions:}
\begin{enumerate}
\item \textbf{Visual Document RAG for Robotics}: We demonstrate that retrieval of visual assembly manuals significantly improves robotic task planning, with Full Manual retrieval achieving +20.4\% F1 improvement over zero-shot baselines.

\item \textbf{Parts Mapping via 3D-to-2D Rendering}: We address the challenge of grounding abstract part identifiers (part\_0, part\_1, etc.) to physical components by rendering 3D part models into labeled 2D overview images, enabling the LLM to bridge symbolic and visual representations.

\item \textbf{Systematic Retrieval Strategy Analysis}: We compare exact document retrieval (BM25) versus similar example retrieval (CLIP-FAISS), finding that access to the exact manual outperforms few-shot learning from similar items (F1: 0.537 vs 0.513).

\item \textbf{End-to-End Simulation}: We demonstrate the complete pipeline from document retrieval to assembly execution in NVIDIA Isaac Sim, showing that retrieval-augmented reasoning can directly drive simulated robot behavior.
\end{enumerate}

\textbf{Key Findings:} Task complexity (number of parts) is the primary performance bottleneck. The model exhibits systematic under-prediction rather than hallucination—predicting 743 connections versus 1,131 ground truth—suggesting that visual procedural understanding, not retrieval quality, is the key challenge for future work.

\textbf{Future Directions:} Our framework opens several research directions: (1) improving visual instruction following for complex multi-part assemblies through specialized architectural components, and (2) extending the framework to other procedural domains such as manufacturing, cooking, and medical procedures where visual instructions guide physical actions.

\bibliographystyle{plain}
\bibliography{custom}



\end{document}